\documentclass{egpubl}
 
\JournalSubmission    
\usepackage[T1]{fontenc}
\usepackage{dfadobe}  

\usepackage{cite}  

\BibtexOrBiblatex
\electronicVersion
\PrintedOrElectronic
\ifpdf \usepackage[pdftex]{graphicx} \pdfcompresslevel=9
\else \usepackage[dvips]{graphicx} \fi

\usepackage{egweblnk} 

\usepackage{amsmath}
\usepackage{booktabs}
\usepackage{color}
\usepackage{xcolor}
\usepackage{color}
\usepackage[ruled]{algorithm2e}
\usepackage{graphicx}
\usepackage{amsmath}
\usepackage{algorithmic}
\usepackage{array}
\usepackage{booktabs}
\usepackage{verbatim} 
\usepackage[english]{babel}
\usepackage{multirow}
\usepackage{amssymb}

\newcommand{\jdq}[1]{{\color{black}#1}}
\newcommand{\YL}[1]{{{\color{black}{#1}}}}
\newcommand{\yl}[1]{\textcolor{black}{{#1}}}

\title[EG \LaTeX\ Author Guidelines]%
      {\LaTeX\ Author Guidelines for EUROGRAPHICS Proceedings Manuscripts}

\author[D. Fellner \& S. Behnke]
{\parbox{\textwidth}{\centering D.\,W. Fellner\thanks{Chairman Eurographics Publications Board}$^{1,2}$\orcid{0000-0001-7756-0901}
        and S. Behnke$^{2}$\orcid{0000-0001-5923-423X} 
        }
        \\
{\parbox{\textwidth}{\centering $^1$TU Darmstadt \& Fraunhofer IGD, Germany\\
         $^2$Graz University of Technology, Institute of Computer Graphics and Knowledge Visualization, Austria
       }
}
}

\title{Reconstructing Recognizable 3D Face Shapes based on 3D Morphable Models}

\author[Diqiong Jiang, Yiwei Jing, Zhang, Fang-Lue, Lai, Yu-Kun, Deng, Risheng, Tong, Ruofeng, Tang, Min]
{\parbox{\textwidth}{\centering Diqiong Jiang$^{1}$, Yiwei Jin$^{1}$,  Fang-Lue Zhang$^{2}$, Yu-Kun Lai $^{3}$, Risheng Deng$^{1}$, Ruofeng Tong$^{1}$, 
        and  Min Tang$^{1}$ 
        }
        \\
{\parbox{\textwidth}{\centering $^1$ Zhejiang University, College of Computer Science
Hangzhou, Zhejiang, CN \\
         $^2$Victoria University of Wellington
Kelburn
Wellington, NZ 6140 \\
         $3$ Cardiff University, School of Computer Science and Informatics
Cardiff, South Glamorgan, UK
       }
}
}

\begin{document}


\maketitle

\begin{abstract}
Many recent works have reconstructed distinctive 3D face shapes by aggregating shape parameters of the same identity and separating those of different people based on parametric models (e.g., 3D morphable models (3DMMs)). However, despite the high accuracy in the face recognition task using these shape parameters, the visual discrimination of face shapes reconstructed from those parameters remains unsatisfactory. Previous works have not answered the following research question: Do discriminative shape parameters guarantee visual discrimination in represented 3D face shapes? This paper analyzes the relationship between shape parameters and reconstructed shape geometry\yl{,} and proposes a novel shape identity-aware regularization (SIR) loss for shape parameters, aiming at increasing discriminability in both the shape parameter and shape geometry domains. Moreover, to cope with the lack of training data containing both landmark and identity annotations, we propose a network structure and an associated training strategy to leverage mixed data containing either identity or landmark labels. \jdq{In addition, since face recognition accuracy does not mean the recognizability of reconstructed face shapes from the shape parameters, we propose the SIR metric to measure the discriminability of face shapes. We compare our method with existing methods in terms of the reconstruction error, visual discriminability, and face recognition accuracy of the shape parameters and SIR metric. Experimental results show that our method outperforms the state-of-the-art methods. The code will be released at \url{https://github.com/a686432/SIR}.}

\end{abstract}



\section{Introduction}

Facial shape estimation from a single RGB image has been an active research topic in both computer vision and computer graphics, and has various applications in fields such as VR/AR, animation, face editing, and biometrics. Early works \cite{blanz1999morphable, blanz2003face,paysan20093d,zhu2015high} focused on 
ensuring the projection of 3D faces is faithful to the input image by minimizing sparse landmark location losses and dense photometric losses. However, if we only use the supervision from the discrepancy between the input image and the projected counterpart, the face shapes reconstructed from different images of the same person may look dissimilar, making them difficult to visually recognize. 
A fundamental reason is that the expression and pose \YL{have much more significant impact on such reconstruction losses than discriminative features for individual subjects.}
\YL{Empirically,} the face shape contributes much less error than the expression and pose in landmark location losses. Therefore, minimizing only the discrepancy between the input image and the projected counterpart makes it difficult to find the optimal face shape \yl{consistently} among different images of the same person. Based on this observation, learning to regress a \emph{recognizable} 3D face shape from a single image with varying poses and expressions has attracted much attention in recent years. A straightforward solution to this problem is aggregating the shape parameters of the same \YL{person} and separating those of different people. Tran et al.~\cite{tuan2017regressing} pool shape parameters belonging to the same \YL{person} to decrease their intraclass variance. Sonyal et al.~\cite{sanyal2019learning}, and Liu et al.\cite{liu2018disentangling} apply shape consistency losses to make shape parameters discriminative and recognizable. Their shape parameters achieve sustainable high performance in face recognition, but the \YL{resulting} 3D face \YL{shapes} still \YL{fail} to be visually discriminative, since the authors focus on improving the discrimination of shape parameters while ignoring the relationship between shape parameters and shape geometries. Therefore, to transfer the discrimination of shape parameters to 3D geometries, the relationship between shape parameters and 3D geometries needs to be carefully investigated\YL{, rather than simply} applying shape consistency losses to shape parameters.

The aim of our research is to reconstruct a stable and recognizable 3D face shape \YL{from an input image}. More \YL{specifically}, the reconstructed \YL{face shapes} from the proposed method must meet the following criteria: (1) the \YL{neutral} face shapes of the same identity must have low error with each other\YL{, based on some geometric metric such as \yl{the} root mean squared error (RMSE).} 
(2) \YL{Neutral face shapes of different people must be sufficiently different (e.g. with high RMSE)} to ensure that the differences can be visually perceived by humans. 
(3) The reconstructed 3D face shapes under different expressions and poses need to be visually identifiable, \YL{as} those reconstructed from a neutral frontal face image. To achieve the above goals, \YL{as we will later show in Sec.~\ref{sec:properties}, } the following conditions need to be satisfied in the method: (1) the 3D shape parameters \yl{need to} be discriminative \YL{under} the Euclidean distance; (2) the centers of 3D shape parameters of the same identity are the parameters regressed from \yl{neutral} frontal face images; (3) the 3D shape basis is orthonormal; and (4) the distribution of the 3D shape parameters follows a particular multivariate Gaussian distribution, which is an inherent property when constructing the 3D morphable model (3DMM).
In this paper, we design a novel shape identity-aware regularization (SIR) loss, which explicitly imposes shape consistency on the shape parameter space and implicitly guides generated face shape geometry to be visually recognizable. As shown in Table \ref{tab:Cond}, the loss functions proposed by existing methods \cite{liu2018disentangling,sanyal2019learning} do not satisfy all these conditions at the same time. 

\begin{figure}[htbp]
\begin{center}

\centering 
\includegraphics[width=9cm,trim={1cm 3cm 13cm 1cm},clip]{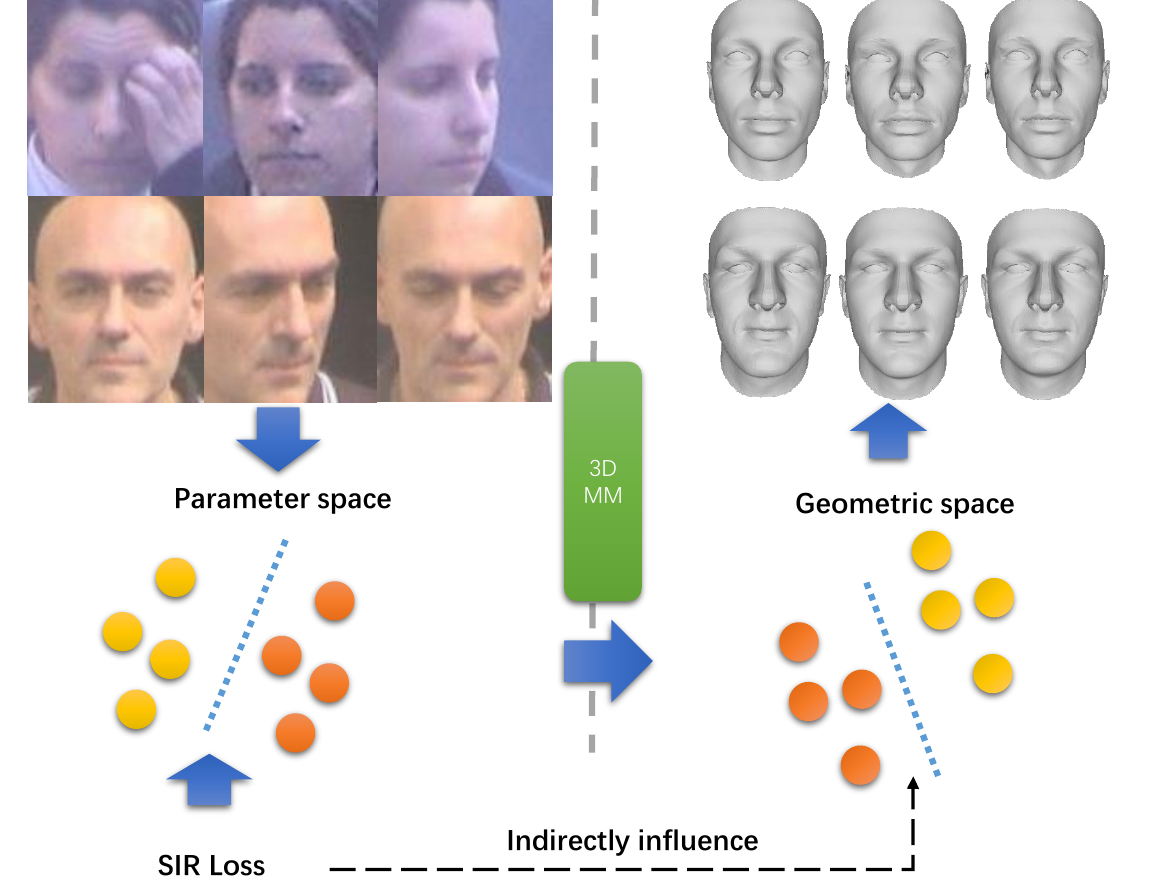} 
\end{center}
\caption{\yl{The} SIR loss is designed according to the relationship between 3DMM parameter
space and geometric space. Therefore, although it \YL{only} imposes shape
consistency directly onto the shape parameters, it essentially guides the face
geometry. These samples show that the represented face shapes are visually discriminative.}
\label{fig:TOP_fig} 
\end{figure}

\YL{Moreover,} the lack of a large database containing both identity information and 3D geometric information (3D face geometry or facial landmarks) also makes the task of learning recognizable face shapes difficult. To obtain a sufficient amount of training data, Tran et al. \cite{tuan2017regressing} and Liu et al. \cite{liu2018disentangling} use 3D face geometries in their methods, which are produced by 3D facial labels estimated from 2D images. Their methods are limited by the capability of geometry reconstruction algorithms. Another work\cite{sanyal2019learning} \YL{relies on detecting facial landmarks by a facial detector}, but the detected \YL{landmark locations} are inaccurate, especially in challenging situations. We apply a more flexible way of dealing with the lack of training data. Our network and training strategy can accept images labeled with \YL{either} identity or geometry information such as facial landmarks, and combine them during the training process. Consequently, the generation of extra \YL{databases} with annotations of both identity and face geometry is unnecessary. In fact, our network allows us to use any off-the-shelf face recognition and face reconstruction database for training. 

\jdq{In addition, \yl{previous} works use the face recognition accuracy of the shape parameters to measure how recognizable their reconstructed face shapes are. However, we observe that this metric cannot fully reflect the face shapes' recognizability. We thus propose a new Shape Identity-aware Regularization (SIR) \yl{metric} to measure the recognizability of face shapes and conduct a user study to prove the relationship between the SIR metric and the discriminability of reconstructed face shapes.}

This paper investigates the relationship between the 3DMM parameter space and 3D geometric space and presents a method to transfer discrimination from the 3DMM parameter space to the geometric space. We propose a novel SIR loss function for face reconstruction\YL{, which comprises} two terms: an identification term, including inter-class separation loss and intraclass aggregation loss, and a parameter distribution term. As Figure \ref{fig:TOP_fig} shows, the SIR loss explicitly imposes shape consistency on shape parameters while implicitly guiding face shapes such that they are visually discriminative. The main contributions of this paper include the following:
\begin{itemize}
\item We investigate the relationship between the 3DMM parameter space and 3D geometric space and propose that a deep model should follow four principles so that the \YL{resulting face shapes} are discriminative in both the parameter and geometry domains. 
\item We propose a deep network that is capable of transferring discriminative features from the shape parameter space to the geometry space with off-the-shelf face recognition and face reconstruction datasets as training data. We also propose an effective training paradigm that leads our network \YL{to} robustly converge with incompletely labeled training data.
\item We propose the SIR loss, which explicitly regularizes 3DMM shape
parameters to satisfy all four aforementioned conditions while implicitly guiding face shapes to be visually discriminative. The parameter distribution term of the SIR loss ensures that the shape geometry discrimination is also visually discriminative. 
\jdq{
\item We propose a new metric which measures the stability of face shape reconstructed from images of the same person and the distinguishability from images of different people.}
\end{itemize}

\begin{table}
\begin{center}
\begin{tabular}{lcccc}
\hline
\toprule
Condition & (1) & (2) & (3) &(4) \\
\midrule

Liu et al.\cite{liu2018disentangling} &  &  & \checkmark &  \\
Ringnet\cite{sanyal2019learning}  &  \checkmark &  & \checkmark & \\
Ours &  \checkmark &  \checkmark &  \checkmark &  \checkmark \\
\bottomrule 
\end{tabular}
\end{center}
\caption{The \YL{conditions} satisfied by different methods. (1) The 3D shape parameters are discriminative with regard to the Euclidean distance; (2) the centers of 3D shape parameters with the same identity are parameters regressed from a neutral frontal face image; (3) the 3D shape basis is orthonormal; (4) the 3D shape \YL{parameters satisfy} a particular multivariate Gaussian distribution.} 
\label{tab:Cond}
\end{table}
\section{Related Work}
In the 3D face reconstruction field, 3D faces are reconstructed from various inputs, including a depth map \cite{kazemi2014real}, video\cite{garrido2016reconstruction,tewari2019fml,cao2014displaced,huber2016multiresolution,saito2016real}, multi-view images\cite{roth2016adaptive,piotraschke2016automated,roth2015unconstrained} and a single image\cite{richardson2017learning,tewari2017mofa,jackson2017large,tewari2018self}. Among them, reconstructing a 3D face from a single image has attracted more attention because of its simplicity and \YL{wide applicability}. After several years of research, monocular reconstruction was generalized from coarse-level reconstruction by parametric face models~\cite{blanz2003face} to medium-level~\cite{li2013realtime,bouaziz2013online} and fine-level shape~\cite{richardson2017learning,yamaguchi2018high} corrections. Recently, some works~\cite{tuan2017regressing,liu2018disentangling,sanyal2019learning} considered using shape consistency to make reconstructed face shapes recognizable. In the rest of this section, we focus on 3DMM face reconstruction and shape-consistent face reconstruction which are more closely related to our work.

\textbf{Monocular 3D face reconstruction based on 3DMM.} The groundbreaking work of monocular 3D face reconstruction with statistical models can be traced back to Blanz and Vetter\cite{blanz1999morphable,blanz2003face} which recovered facial geometries by solving an optimization problem constrained by linear statistical \YL{models}, i.e., 3DMMs. Paysan et al. \cite{paysan20093d} and Zhu et al. \cite{zhu2015high} extend the 3DMM with pose and expression parameters. In recent years, Deep \yl{Convolutional Neural Networks} (DCNNs) have shown strong capabilities in many computer vision tasks. The existing literature \cite{dou2017end,zhu2016face,tuan2017regressing,tewari2018self,guo2018cnn,richardson2017learning} reveals that CNNs can effectively regress the 3DMM parameters with sufficient training data. They provide comparable reconstruction precision with much less computation \YL{time} and adapt to input images under challenging conditions.  Richardson et al.\cite{richardson20163d} build a synthetic dataset using the 3DMM with random shape, expression, and pose parameters, and render them as 2D images with different levels of illumination. However, the synthesized data cannot capture the complexity of the real world. Zhu et al.\cite{zhu2017face} fit 3D shapes with traditional methods and augment data by applying the image warping technique to simulate in-plane and out-of-plane head rotation. They build the 300W-LP dataset, which covers various head poses and facial expressions with labeled 3DMM coefficients. In this way, the shape labels are ambiguous and inaccurate because they are constrained only by sparse facial landmarks in the fitting process. Sanyal et al.\cite{sanyal2019learning} regress 3D shape parameters without any supervised 2D-to-3D training data. The landmark labels are detected by a face detection algorithm, which \YL{are} not very precise in challenging conditions (e.g., large poses and poor lighting conditions). \YL{Furthermore,} sparse landmarks cannot capture sufficient recognizable features in face geometries. With the development of generic differentiable rendering\cite{zhu2020reda,liu2019soft,kato2018neural}, \cite{tewari2017mofa,deng2019accurate,tran2019towards} train networks without shape labels in an unsupervised or weakly-supervised way by constraining the consistency between \YL{rendered and input images}. 
The photometric consistency can capture more geometric details, especially from the frontal face. Our method uses the 300W-LP facial landmarks and the pixelwise photometric difference as our reconstruction training losses.

\textbf{Shape-consistent face reconstruction.}
Many works \cite{zhu2016face,jackson2017large,feng2018joint} pursue alignment accuracy or pixelwise appearance accuracy to get precise face geometries. However, the final face geometry is composed of face shape, expression and pose.
Any of those parameters could dominate the reconstruction if the model is poorly trained. Therefore, a well-aligned face geometry does not guarantee the accuracy of a face shape. To reconstruct a stable and visually discriminative face shape, Tran et al. \cite{tuan2017regressing} label a \YL{large} number of face images with 3DMM shape parameters and develop a deep CNN to learn the mapping from images to shape parameters. During training, they pool coefficients that belong to the same identity to give their output \YL{features} lower intraclass variance. Liu et al.~\cite{liu2018disentangling} propose a \YL{multi-task} deep CNN to disentangle identity from ``residual attribute'' to learn the 3D face shape and discriminative authentication feature together. They use a softmax loss function to directly push away the shape parameters of different people while aggregating those of the same person. Compared with pooling, their loss function can achieve even better face recognition accuracy with a simpler network structure. Sanyal et al.\cite{sanyal2019learning} achieve a similar goal by introducing a shape consistency loss embodied in a ring-structured network. \YL{However, their method only aims} to achieve shape parameter consistency rather than \YL{visual consistency of shape geometry}, and does not attempt to explain the relationship between shape parameter consistency and visual shape geometric consistency. Our work takes both shape \yl{parameters} and geometry discrimination into consideration and proposes SIR loss to separate shape parameters explicitly and distinguish face geometries implicitly. 
\section{Our Method}
This section first introduces the parametric face model. Then, we investigate the relationship between the 3DMM parameter space and 3D geometry space and propose principles that the deep neural network should follow to make the results discriminative in both the parameter and geometry domains. Finally, the network, loss function, and training strategy are designed to make our deep neural network satisfy these principles.
\subsection{Parametric Face Model}
We follow the previous work\cite{zhu2016face} which combines the Basel Face Model-09 \cite{paysan20093d} and FaceWarehouse \cite{cao2013facewarehouse} by Equation(\ref{3dmm}) for our 3DMM representation to describe the geometry of a 3D face model
\begin{equation}
\mathbf{S} = \mathbf{\bar{S}+A}_{id}\alpha_{id}+\mathbf{A}_{exp}\alpha_{exp}
\label{3dmm}
\end{equation}
where $\mathbf{S}\in \mathbb{R}^{3n}$ is a reconstructed 3D face with $n$ vertices, which is controlled by the shape parameter vector $\alpha_{id}$ and the expression parameter vector $\alpha_{exp}$ for representing various shape identities and expressions. $\mathbf{\bar{S}}\in \mathbb{R}^{3n}$ is the mean face shape. The orthogonal matrices \(\mathbf{A}_{id}\) and \(\mathbf{A}_{exp}\) are the bases of shape and expression, respectively. 

Six degrees-of-freedom (rotation and translation) are required to describe the camera pose. More specifically, 3DMM meshes are transformed by the camera pose \([\mathbf{R}|\mathbf{t}_{3d}]\in SE3\) by the following Equation(\ref{3dmm_pose})
\begin{equation}
\mathbf{V}_{3d} = \mathbf{R\cdot
(\bar{S}+A}_{id}\alpha_{id}+\mathbf{A}_{exp}\alpha_{exp})+\mathbf{t}_{3d}
\label{3dmm_pose}
\end{equation}
where \(\mathbf{V}_{3d}\) denotes the 3D vertices of the transformed 
\YL{3DMM mesh} in the camera coordinate system. $\mathbf{t}_{3d}\in \mathbb{R}^{3n}$ is the translation matrix and we use a rotation matrix \(\mathbf{R}\) converted from a quaternion to represent rotation.

We apply weak projections to project the 3DMM meshes to the image plane so a scalar \(f\) \YL{is} introduced as the focal length to perform the projection as in Equation(\ref{3dmm_pose_proj}).
\begin{equation}
\mathbf{V}_{2d} \!=\! f\cdot\mathbf{Pr\cdot
R\cdot(\bar{S}\!+\!A}_{id}\alpha_{id}\!+\!\mathbf{A}_{exp}\alpha_{exp})\!+\!\mathbf{t}_{2d}
\label{3dmm_pose_proj}
\end{equation}
where $\mathbf{V}_{2d}$ denotes the projected 2D coordinates of the 3D model\YL{,  and $\mathbf{Pr}$ is the projection matrix} $\bigl(\begin{smallmatrix}1 & 0 & 0\\ 0 & 1 & 0\end{smallmatrix}\bigr)$.

\subsection{Properties of 3DMM model}\label{sec:properties}

\begin{figure*}[ht]
\begin{center}   
\centering 
\includegraphics[width=17cm]{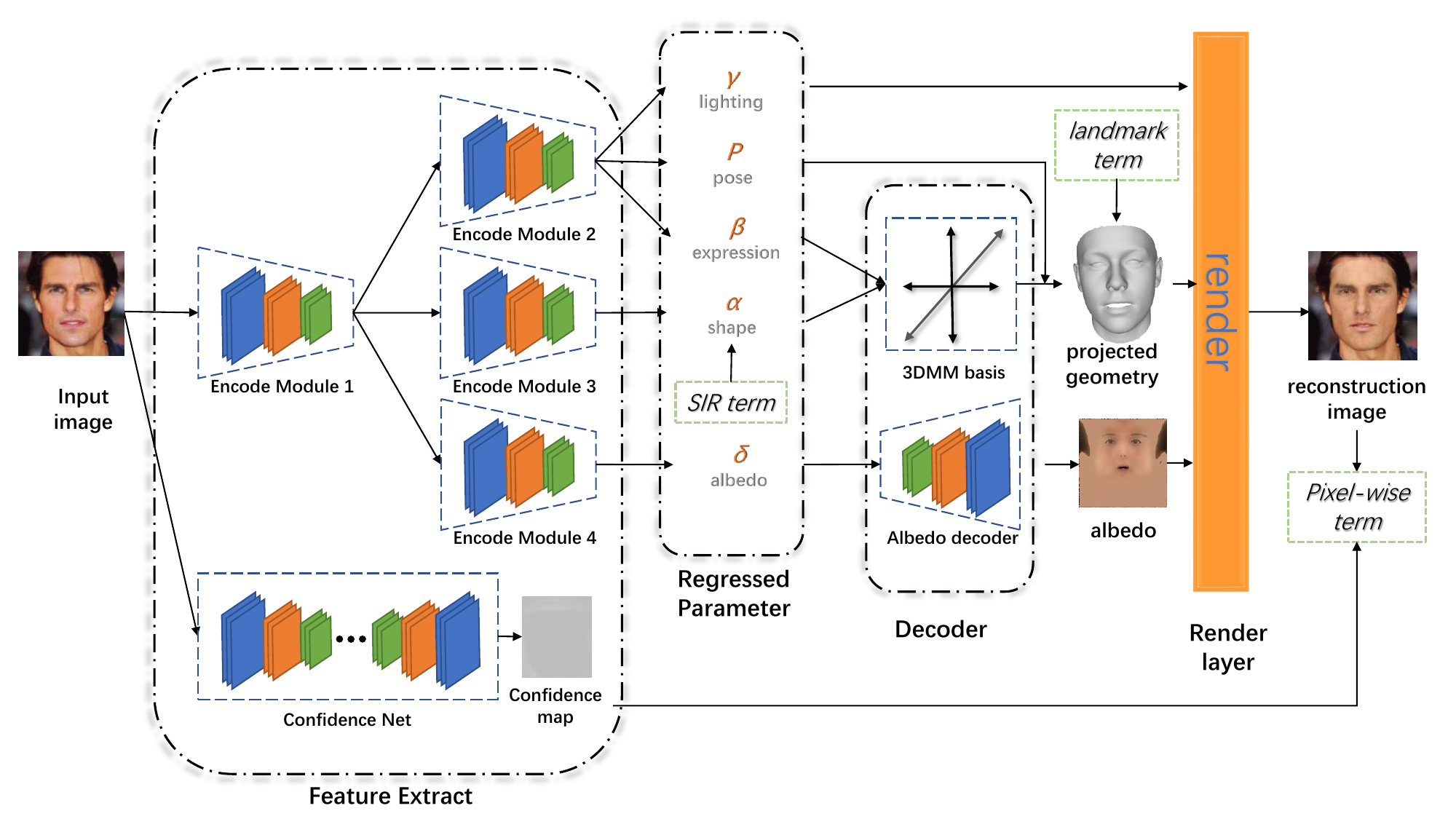} 

\end{center}   
\caption{\textbf{The framework of our method.} Our network contains a feature extraction module (Encoder Module 1) followed by three encoders (Encoder Modules 2, 3 and 4), \jdq{which regress different face parameters for rendering respectively from the same output of Encoder Module 1. The confidence map gives the probability of each pixel belonging to skin, which provides pixel-wise weights for the pixel loss and the perceptual loss. In addition, by enabling the SIR term and the landmark term}, even incompletely labeled data can effectively train the network.}
\label{fig:pipeline} 
\end{figure*}

In this subsection, we explore the underlying relationship between 3DMM shape parameters and 3D shape geometries \yl{as well as} the conditions in which the separable shape parameters lead to visually distinguishable face geometries. For simplicity, we focus only on face shapes regardless of expression and pose in this \YL{subsection}, \YL{so a simplified 3DMM model is used in the following discussion.}

\begin{figure}[h]
\begin{center}   
\centering 
\includegraphics[width=9cm,trim={1.5cm 4cm 8cm 2.5cm},clip]{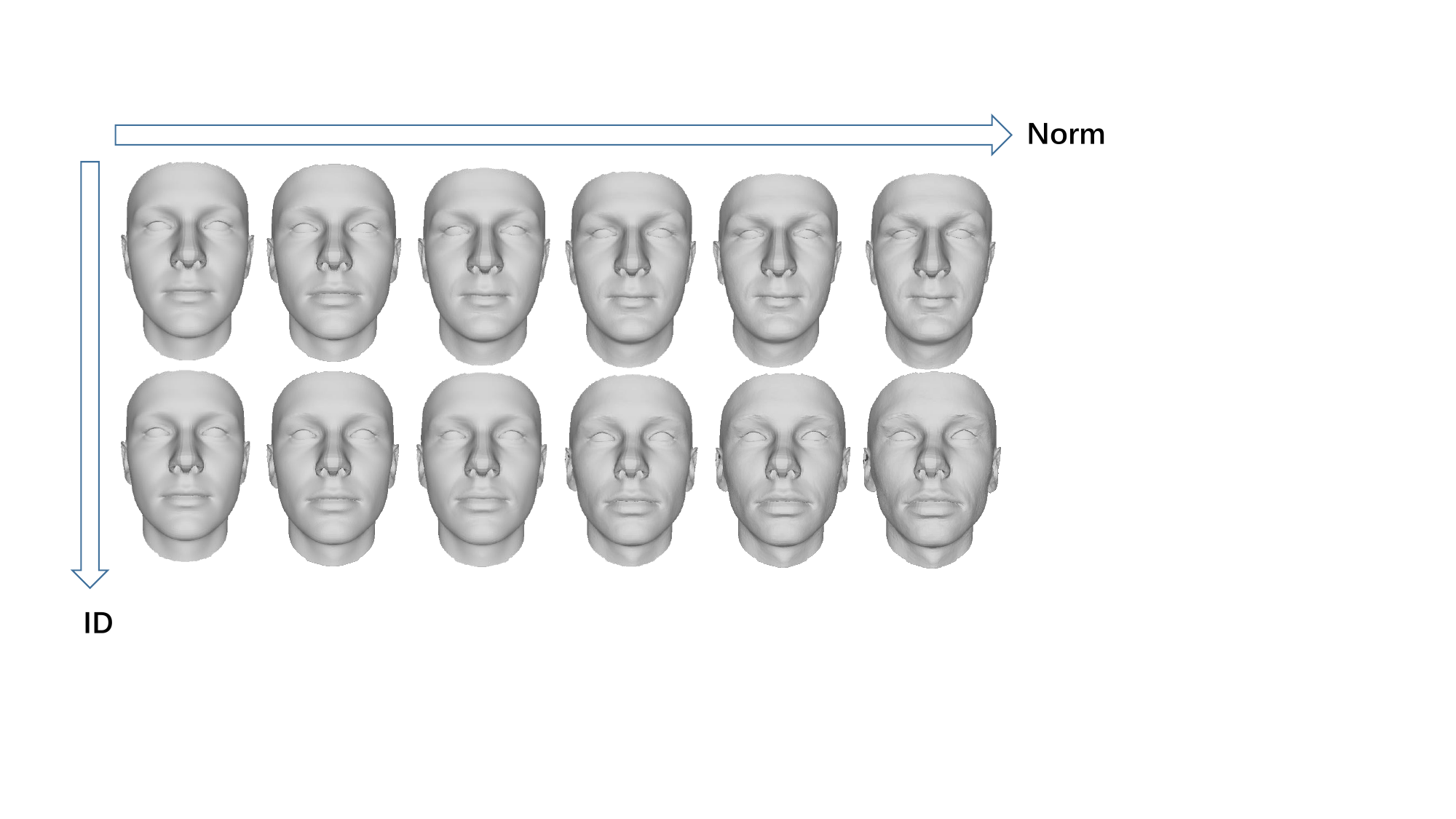} 
\end{center}
\caption{From left to right in each row, the shape \YL{parameters are} multiplied by 0.1, 0.4, 0.6, 0.8, 1.0 and 1.2. The face shapes in different rows represent different people.}
\label{fig:Norm}
\end{figure}

\textbf{From parameter discrimination to shape geometry discrimination.}
We denote $A \in \mathbb{R}^{3n \times m}$ as the shape basis, $\alpha \in \mathbb{R}^m$ as the shape parameters and $\mathbf{\bar{S}}\in \mathbb{R}^{3n}$ as the mean face \YL{shape}. $m$ is the dimension of the shape parameters, and $n$ is the number of vertices of the face shape. The face shape $\mathbf{S}$ is represented by Equation(\ref{3dmm_shape}). 
\begin{equation}
\mathbf{S} = \mathbf{\bar{S}+A\alpha}
\label{3dmm_shape}
\end{equation}
Suppose $\mathbf{x}$, $\mathbf{y} \in \mathbb{R}^m$ are the shape parameters of two faces. Denote $e=\frac{1}{m}\left \| \mathbf{x}-\mathbf{y} \right\|_2^2$ as the square of their Euclidean distance. Accordingly, $\mathbf{X}$, $\mathbf{Y}\in \mathbb{R}^{3n}$ are their corresponding face shapes and the square of their Euclidean distance is $E=\frac{1}{n}\left \| \mathbf{X}-\mathbf{Y} \right \|_2^2$. According to Equation(\ref{3dmm_shape}), $E$ can be calculated by:
\begin{equation}
E =\frac{1}{n} \left \| \mathbf{X}-\mathbf{Y} \right \|_2^2 = \frac{1}{n}\left \| \mathbf{A}(\mathbf{x}-\mathbf{y})
\right \|_2^2
\label{shape_error}
\end{equation}

\YL{Assuming} that $\mathbf{A}$ is \yl{an} orthonormal basis, the relationship between $E$ and $e$ is as shown in Equation(\ref{relationship_error}), which means that geometric Euclidean distance is proportional to the parameter Euclidean distance.
\begin{equation}
E = \frac{1}{n}\left \| \mathbf{x}-\mathbf{y} \right \|_2^2 = \frac{m}{n} e 
\label{relationship_error}
\end{equation}
This equation proves that when we minimize (maximize) the Euclidean distance between the parameters of face shapes of the same person (different \YL{persons}), their corresponding geometric distances are smaller (larger) as well, which suggests that it is feasible to formulate recognition errors using the Euclidean distance (e.g. center loss functions) in addition to the cosine distance (e.g. softmax-like loss functions).

\textbf{From shape geometry discrimination to visual discrimination.} As mentioned above, in Euclidean space, separation of the shape parameters ensures separation of the shape geometries. However, even though the shape geometries can be separated numerically, we cannot ensure that the separation is visually recognizable, since people usually fail to perceive small differences between meshes. As shown in \yl{Figure~}\ref{fig:Norm}, in each column, shape parameters are multiplied by \YL{a different factor}. \YL{When shape parameters of a set of faces are scaled by a factor, it does} not influence the separation of parameters and geometries numerically. 
However, even though the shapes \YL{in each column can be equally separated numerically through a classifier}, \YL{visually} we find it difficult to distinguish face shapes when the parameter norms are relatively small, implying that the same geometry discrimination could have various degrees of visual discrimination. 
We find that \YL{only with} the center loss and softmax-like loss on shape parameters, 
the network has a high probability of falling into a local minimum \YL{that resembles} an average face, \YL{where} the norm of regressed shape parameters is minimal. This observation suggests that in addition to the individual shape geometry, we also need to pay attention to adding additional constraints to make the shape geometries visually distinguishable. According to \YL{Equations (\ref{3dmm_shape}) and (\ref{relationship_error})}, the norm of the shape \YL{parameters} $ \left \| \alpha \right \|_2$ is proportional to  $ \left \|  \mathbf{A}\alpha \right \|_2$, \YL{i.e.,} the residual between facial geometry $S$ and mean face $\bar{S}$. Therefore, by constraining the parameters to fit an appropriate distribution, the geometric residuals of faces can be sufficiently large to make the shapes intuitively distinguishable. 

The 3DMM is based on assuming that the data (face \YL{vertex positions}) \YL{follows} a multivariate normal distribution. This assumption is a necessary prerequisite of principal component analysis (PCA). The multivariate normal distribution of shape parameters is given by\cite{blanz1999morphable}:
\begin{equation}
p(\alpha)\sim \exp[-\frac{1}{2}\sum_{i=1}^{m}(\alpha_i/\sigma_i)^2]
\label{Eq:distribution}
\end{equation}
where $\sigma_i$ is the \yl{$i$-th eigenvalue} of the shape covariance matrix. If our trained model outputs shape parameters that have the same distribution as in Equation(\ref{Eq:distribution}), the reconstructed face shape should share the same level of visual distinctiveness \YL{as} the scan data used to build the 3DMM. According to Equation(\ref{Eq:distribution}), the parameters divided by the \YL{eigenvalues follow} the standard normal distribution. \YL{Therefore, we propose to} minimize the KL divergence between the parameters divided by the eigenvalues and the standard normal distribution to constrain the \yl{parameters} to fit the distribution described in Equation(\ref{Eq:distribution}).
\begin{equation}
\underset{\theta}{\operatorname{argmin}} KL(\boldsymbol{P}(\boldsymbol{\alpha}
/ \boldsymbol{\sigma} \mid \boldsymbol{I}, \boldsymbol{\theta}) \|
\boldsymbol{N}(0,1))
\label{Eq:Kl_dis}
\end{equation}
where $I$ is the input image and $\theta$ is the weight of the network. Once the shape parameters fit a multivariate Gaussian distribution, the norm is sufficiently large to represent visually discriminative face shapes.

\textbf{From parameter discrimination to visual discrimination.}
To reconstruct a visually recognizable face shape, the reconstructed 3D face shapes from images with different lighting, expressions and poses should be the same as the one from the neutral frontal face image of the same person. \YL{So ideally} in the parameter domain, shape parameters of the same identity \YL{should be} tightly grouped around the \YL{parameters} regressed from the neutral frontal face image. To meet this condition, we modify the center loss~\cite{wen2016discriminative} to push shape parameters of the same identity towards the center of its class and update the center by assigning higher weights to the samples with smaller expression and pose variance. Therefore, the parameters regressed from the neutral frontal face image have a more significant impact on the class center, and the shape parameters from \YL{non-neutral} and \YL{non-frontal} faces will approach those from the neutral frontal face using the modified center loss. In summary, to transfer parameter discrimination to visual discrimination, the shape parameters should satisfy the following conditions: (1) the shape parameters are discriminative in Euclidean space; (2) shape parameters follow a specific multivariate Gaussian distribution; and (3) the centers of the shape parameters are the parameters regressed from the neutral face image of the identity.

\subsection{Network Structure}
Our network has four branches to regress the albedo parameters, shape parameters,  \YL{a} confidence map and other non-identity information (expression parameters, camera parameters and illumination parameters). We use the same residual block (Sphere64a) used in SphereFace\cite{liu2017sphereface}. As shown in Figure \ref{fig:pipeline}, Encode \YL{Modules} 2, 3 and 4 contain last two blocks (Conv3.x,Conv4.x) of Sphere64a, followed \YL{by a fully-connected (FC) layer}.
They share the weights of Encode Module 1, which contains the first two blocks (Conv0.x,Conv1.x) of Sphere64a. Encode Module 1 serves as the feature extraction module, \YL{while} Encode Modules 2-4 serve as the feature separation modules. The feature extraction module shares its low-level features with the feature separation modules, reducing the number of parameters of the whole network. Encode Modules 2-4 improve the network ability to separate high-level features. By balancing the depth of the feature extraction module and that of the feature separation \YL{modules}, the network can obtain better results with fewer network parameters. The \YL{FC layer} adapts the output features to the size of our parameters (199-dim of the shape parameter, 29-dim of the expression parameters, 7-dim of the camera parameters, 27-dim of the illumination parameters and 512-dim of the albedo parameters). We use the same ConfNet structure to generate \YL{a} confidence map as in \cite{wu2020unsupervised}. The decoder of the albedo consists of a transposed convolution network and regresses the albedo UV map with a resolution of 256$\times$256. We regard the 3DMM basis as \YL{a fully-connected layer with fixed connection weights} 
in the neural network. We build the rendering layer based on Pytorch3d implementation, and the illumination model is a spherical harmonic illumination model.

\jdq{\subsection{Loss function}\label{subSection:loss}
\yl{As previously described, our method can be trained using existing datasets that only contain part of required labels, including face recognition datasets which contain identity labels associated with input images, but without ground truth 3D reconstruction, and 3D face reconstruction datasets which contain 3D reconstruction but without identity labels. To address this, our}
loss function contains three terms: a landmark term, a pixelwise photometric term and a SIR regularization term. According to the existing labels of the training samples, we determine which terms can take effect. For example, if a training sample has \yl{the identity label}, the SIR term and pixelwise photometric term will be enabled. Otherwise, the face reconstruction term and pixelwise photometric term will take effect. All $\varepsilon_*$'s mentioned in this subsection represent balancing weights of the loss function.
\begin{equation}
L =\left\{
\begin{array}{lcl}
 \varepsilon_{l} L_{land} + L_{pixel}      &      & {I \in
S_{recon}}\\
 \varepsilon_{s} L_{SIR} + L_{pixel}       &      & {I \in  S_{id}}
\end{array} \right.
\label{loss_overall}
\end{equation}
In the above equation, \(L_{land}\) denotes the landmark term, \(L_{SIR}\) denotes the SIR term and \(L_{pix}\) denotes the pixelwise photometric term. $S_{recon}$ is the face reconstruction dataset, $S_{id}$ is the face recognition dataset and $I$ is the input image. In the rest of this section, we explain the three loss terms in detail. 
}

\textbf{Landmark term.}
The landmark term $L_{proj}$ simply uses the $L_2$ loss between projected landmarks $\hat{V}_{2d}$ and ground-truth landmarks ${V}_{2d}$. 
\begin{equation}
L_{land} = \frac{1}{N} \left \| V_{2d}-\hat{V}_{2d} \right \|_2
\label{loss_proj}    
\end{equation}
where $N$ is the number of landmarks.

\jdq{
\textbf{Pixelwise photometric term.}
As shown in Equation(\ref{loss_recon}), the pixelwise term consists of two losses: $L_{recon}$ and $L_{reg}$.  
\begin{equation}
L_{pixel} = L_{recon} +  \varepsilon_{reg} L_{reg}
\label{loss_recon}    
\end{equation}
$L_{recon}$ measures the reconstruction errors by both the pixel loss $L_p$ and perceptual loss $L_{pr}$ on the confidence map\cite{wu2020unsupervised}:
\begin{equation}
	L_{recon} = L_{p}+\varepsilon_{pr} L_{pr}
\label{loss_regular}
\end{equation}
}
The confidence map aims to achieve robustness when occlusions and other challenging appearance variations exist such as beard and hair. The pixel loss is defined as follows:
\begin{equation}
L_{p}(\hat{\mathbf{I}}, \mathbf{I}, \sigma)=-\frac{1}{|\Omega|} \sum_{u v \in \Omega} \ln \frac{1}{\sqrt{2} \sigma_{u v}} \exp -\frac{\sqrt{2} \ell_{1, u v}}{\sigma_{u v}}
\label{pixelLoss}
\end{equation}
where $\ell_{1, u v}=\left|\hat{\mathbf{I}}_{u v}-\mathbf{I}_{u v}\right|$ is the $L_1$ distance between the intensity of input image $\mathbf{I}$ and the reconstruction image $\hat{\mathbf{I}}$ at location (u, v) and $\sigma \in \mathbb{R}_{+}^{W \times H}$ is the confidence map. The perceptual loss mitigates the blurriness in the reconstruction result, which is defined as follows:
\begin{equation} 
L_{pr}^{(k)}\left(\hat{\mathbf{I}}, \mathbf{I}, \sigma^{(k)}\right)=-\frac{1}{\left|\Omega_{k}\right|} \sum_{u v \in \Omega_{k}} \ln \frac{1}{\sqrt{2 \pi}\sigma_{u v}^{(k)}} \exp{-\frac{\left(\ell_{u v}^{(k)}\right)^{2}}{2\left(\sigma_{u v}^{(k)}\right)^{2}}}
\label{perceptual_loss}
\end{equation}
where $\ell_{u v}^{(k)}=\left|e_{u v}^{(k)}(\hat{\mathbf{I}})-e_{u v}^{(k)}(\mathbf{I})\right|$ is the $L_1$ distance between feature maps of the $k$-th layer. $e^{(k)}(\mathbf{I}) \in \mathbb{R}^{C_{k} \times W_{k} \times H_{k}}$ is the $k$-th layer of an off-the-shelf image encoder \yl{$\mathcal{E}$}
(VGG16\cite{simonyan2014very} \YL{is used}) and $\Omega_{k}=\left\{0, \ldots, W_{k}-1\right\} \times\left\{0, \ldots, H_{k}-1\right\}$ is the corresponding spatial domain.  $\sigma^{(k)}$ is a confidence map of perceptual loss. 

$L_{reg}$ avoids overfitting when predicting 3DMM parameters and albedo UV maps, which is defined \YL{as}:
\begin{equation}
	L_{reg} = L_{regp}+\varepsilon_{rega} L_{rega}
\label{loss_regular}
\end{equation}
The regularization term of $L_{regp}$ for 3DMM coefficients is:
\begin{equation} 
L_{regp} = \varepsilon_{id}\sum^{m_{id}}_{j=1}\frac{\alpha_{id_j}^2}{\sigma_{id_j}^2}+ \varepsilon_{exp}\sum^{m_{exp}}_{j=1}\frac{\alpha_{exp_j}^2}{\sigma_{exp_j}^2}
\label{para_regular}
\end{equation}
where $\sigma_{id}$ is the eigenvalue \YL{vector} of the shape basis and $\sigma_{exp}$ is the eigenvalue \YL{vector} of the expression basis.  $\alpha_{id}$ is the shape \YL{parameters} and $\alpha_{exp}$ is expression \YL{parameters}. $m_{id}$ and $m_{exp}$ are the dimensions of the shape and expression parameters respectively.
The regularization of albedo UV maps consists of smooth and residual terms, which penalize differences \YL{between} neighboring pixels and enforce a prior distribution towards the mean albedo to avoid the regressed albedo being \yl{too much} away from the mean albedo.
\begin{equation}
\begin{aligned}
L_{rega}(\mathbf{A})=& \sum_{\mathbf{p}_{i}^{\mathrm{uv}} \in \mathbf{A}^{\mathrm{uv}}}\left\|\mathbf{A}^{\mathrm{uv}}\left(\mathbf{p}_{i}^{\mathrm{uv}}\right)-\frac{1}{\left|\mathcal{N}_{i}\right|} \sum_{\mathbf{p}_{j}^{\mathrm{uv}} \in \mathcal{N}_{i}} \mathbf{A}^{\mathrm{uv}}\left(\mathbf{p}_{j}^{\mathrm{uv}}\right)\right\|_{2} \\
& +\varepsilon_{uv} \left \| \mathbf{A}^{\mathrm{uv}} \right \|_2^2
\end{aligned}
\label{albedo_regular}
\end{equation}
where $\mathbf{A}^{\mathrm{uv}}$ is the albedo UV map and $\mathcal{N}_{i}$ denotes \YL{the} set of 4-pixel neighborhood of pixel $\mathbf{p}_{i}^{\mathrm{uv}}$.

\textbf{Shape identity-aware regularization term.} 
\yl{Our proposed} SIR term includes two components as shown in Equation\eqref{loss_recog}, an identification loss and a Kullback-Leibler Loss.
\begin{equation}
L_{id} =  L_{recog} +   \varepsilon_{kl} L_{kl}
\label{loss_recog}
\end{equation}

To ensure the criteria that shape parameters are discriminative in Euclidean space, the identification loss is defined as in Equation(\ref{Eq:Lrecog}). It combines softmax-like loss and center loss\cite{wen2016discriminative}.
\begin{equation}
L_{recog} =  L_{sm} + \varepsilon_{c} L_{c}
\label{Eq:Lrecog}
\end{equation}
$L_{sm}$ is a softmax-like loss (e.g. softmax, Cosloss\cite{wang2018cosface}, A-softmax\cite{liu2017sphereface}, and Arcloss\cite{deng2019arcface}), which separates parameters and speeds up the convergence and $L_{c}$ discriminates features in Euclidean space (e.g triplet loss and center loss). We choose Cosloss \cite{wang2018cosface} as our softmax-like loss.

To ensure the condition that the centers of shape parameters of the same identity are the parameters regressed from neutral frontal face images, we first calculate the confidence, which indicates similarity to the neutral frontal face.
\begin{equation}
\begin{aligned}
\yl{\hat{f}} =\frac{1}{8}(cos\alpha+1)(cos\beta+1)(cos\gamma+1) \cdot \exp^{-\lambda \left \| \alpha_{exp} \right \|_2}
\end{aligned}
\label{crecog}
\end{equation}
where $\alpha_{exp}$ represents the expression parameters. $\alpha$, $\beta$ and $\gamma$ are Euler angles of 3D face poses.

We use the following formula to update the centers. It assigns higher weights \YL{to} the neutral frontal face.
\begin{equation}
\begin{aligned}
\Delta c_j &= 
\frac{\sum^{\YL{n_b}}_{i=1}\delta(y_i=j)\cdot(c_j-x_{{id}_i})}{1+\sum_{i=1}^{\YL{n_b}}\delta(y_i=j)} \cdot \yl{\hat{f}}
\end{aligned}
\label{Lsdrecog}
\end{equation}
where \YL{$n_b$} is the number of samples in a mini-batch, $\delta(\cdot) = 1$ if the condition is true and $\delta(\cdot) = 0$ otherwise. $y_i$ is the identity label of the sample, $c_j$ is the shape parameter center of the $j$-th class, and $\alpha_{id}$ represents the shape parameters.

To ensure that the shape \YL{parameters satisfy} a specific distribution, we use the Kullback-Leibler loss to constrain the shape parameters to fit a zero-mean  multivariate Gaussian distribution with the eigenvalues as its variances.
\begin{equation}
L_{kl}=KL(\boldsymbol{P}(\boldsymbol{\alpha}
/ \boldsymbol{\sigma} \mid \boldsymbol{I}, \boldsymbol{\theta}) \|
\boldsymbol{N}(0,1))
\label{Eq:Kl_loss}
\end{equation}
\subsection{Training strategy}
\YL{Public} databases usually contain either identity labels or landmark labels. Some existing works need to use face detectors or optimization-based methods to generate the annotation needed for face reconstruction. However, these \YL{annotations are} unsatisfactory in challenging examples\YL{, which} limits the performance of models by those algorithms. Based on the above considerations, we choose to build a new dataset with a mixture of face recognition and facial landmark data. Directly training our network on this mixed dataset with \YL{different types of} labels results in tricky convergence for the following reasons: (1) The numbers of samples of face recognition and face alignment are unbalanced. (2) Incomplete labels can result in an oscillating learning process. (3) The objective function is complicated, making our network easily fall into a local \YL{minimum} without good initialization. Therefore, \YL{it is} important to warm up the network and maintain a balanced proportion of face recognition and face reconstruction data in the mixed database. The warming-up stage consists of two steps. First, we train our network on the 300W-LP~\cite{zhu2015high} database without \YL{the} SIR loss. Second, we train the whole network on the mixed database \YL{with} the SIR loss \YL{added}. The mixed database consists of VGGFace2~\cite{cao2018vggface2} and the 300W-LP~\cite{zhu2015high}. VGGface2 contains 3.31 million images of 9131 subjects covering a large range of poses, ages and ethnicities. 300W-LP is a synthetically generated dataset based on the 300-W database~\cite{sagonas2016300} containing 61,255 samples across various poses. We \YL{only use} the 300W-LP landmark \YL{labels} because the synthetic face \YL{shapes are} not precise. Considering that the sample numbers of the two databases are extremely unbalanced, we design a sampling scheme in which the probability of selecting samples from the face recognition database is given by:
\begin{equation}
P = \frac{N_{recon}}{N_{recog}+N_{recon}}
\end{equation}
where $N_{recon}$ is the number of samples in the face reconstruction dataset and $N_{recog}$ is that in the face recognition dataset. The probability of selecting samples from the face reconstruction database is $1-P$. \jdq{We train our model on a GTX2080Ti GPU with a learning rate of 5e-5 and a batch size of 8.} 
\begin{figure}[h]
\begin{center}   
\centering 
\includegraphics[width=7cm,trim={0cm 0cm 0cm 0},clip]{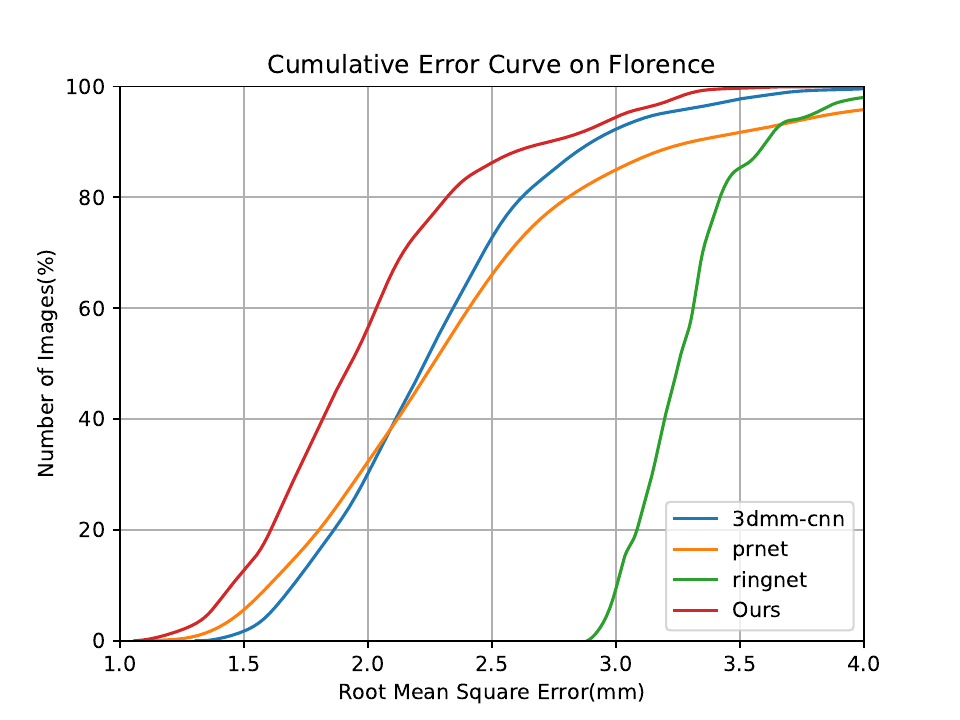} 
\end{center}
\caption{Cumulative error distribution (CED) \yl{curves} on Florence dataset.  We compare our method with Tran et al.~\cite{tuan2017regressing},  Ringnet~\cite{sanyal2019learning} and MGCNet~with\cite{shang2020self}.}
\label{fig:CED_fig} 
\end{figure}
\section{Experimental Results}

\jdq{To evaluate the effectiveness of our method, we measure the face recognition accuracy on the reconstructed faces using the generated shape parameters and test face reconstruction errors on the MICC dataset as the previous works. However, as mentioned above, the face recognition accuracy metric can not fully reflect the recognizability of the face geometries. Therefore, we propose the novel metric named SIR-scores to evaluate the recognizability of face shape. Finally, we qualitatively evaluate the visualized reconstruction results.}


\subsection{Face Recognition Performance}
We design an ablation study to investigate the impact of SIR losses on face recognition performance and compare it with other methods based on the 3DMM models.  First, we introduce the test datasets and evaluation method.

\textbf{Testing benchmarks.} We use the following datasets: (1) LFW\cite{huang2008labeled}, a standard face verification testing dataset. It contains 13,233 labeled face images for 5,749 different individuals with a total of \YL{6,000} defined pairs. (2) CFP, the Celebrities Frontal-Profile dataset~\cite{sengupta2016frontal}. It is aimed at evaluating face identification with frontal and profile pairs and has approximately 7,000 pairs of matches defined by 3,500 same pairs and 3,500 not-same pairs for approximately 500 different subjects. (3) YTF, Youtube face dataset~\cite{wolf2011face}. It contains 3,425 videos of 1,595 individuals. We follow the verification protocol and report the result on \YL{5,000} video pairs.


\textbf{Evaluation method.}
In the methods designed for the face recognition task, the identity of a subject in an image can be represented as \yl{a} learned latent \yl{code}. The similarity between two identity representations (usually based on the cosine distance or Euclidean distance) is calculated to determine whether the images are of the same person. In our evaluation, the shape parameters can be used for identity representation, similar to the latent codes in other face recognition methods.
This evaluation method is not suitable for evaluating shape parameter discrimination 
because the Euclidean distance between parameters can reflect the separation of the face geometry while the cosine distance cannot. Therefore, we directly use the Euclidean distance between 3DMM shape parameters to measure the similarity between two faces. \yl{For fairness,} we also show the results of using the cosine distance.

\begin{table}
\centering

\begin{tabular}{cccccc}
\toprule
\multicolumn{3}{c}{Losses} & \multirow{2}{*}{LFW}  & \multirow{2}{*}{CFP-FP} &\multirow{2}{*}{YTF}  \\

$L_{sm}$ & $L_{c}$ & $L_{wc}$ &     &  & \\
\midrule 
& & & 67.67  & 54.16 & 66.46 \\
$ \checkmark$ & & & 90.55  & 70.77 & 81.48 \\
$ \checkmark$ &  $ \checkmark$ & & \textbf{95.23}  & \textbf{83.45} & \textbf{89.10} \\
$ \checkmark$ & &  $ \checkmark$ & 94.47  & 80.73 & 86.40 \\
\bottomrule 
\end{tabular}
\caption{The face verification accuracy (\%) on LFW, CFP-FP and YTF \YL{for} different losses of the shape identity-aware regularization term. $L_{sm}$ is the cosloss\cite{wang2018cosface}. $L_c$ is the center loss\cite{wen2016discriminative}, and $L_{mc}$ is the weighted center loss.}
\label{tab:FG_acc}
\end{table}
\begin{table}
\begin{center}
\begin{tabular}{|l|c|c|c|}
\hline
Method & LFW & CFP-FP & YTF    \\
\hline
\multicolumn{4}{|c|}{Cosine similarity} \\
\hline
3DMM-CNN  & 90.53  & - & 88.28 \\
Lui et al. & 94.40 & - & 88.74 \\ 
D3FR &88.98 & 66.58 & 81.00 \\
TDDFA & 64.90 & 57.57 & 58.50 \\
MGCNet &82.10 & 70.87 & 75.58 \\
RingNet &79.40 & 71.41 & 71.02 \\
Ours & \textbf{95.36} & \textbf{83.34} &\textbf{89.07} \\
\hline
\multicolumn{4}{|c|}{Euclidean similarity} \\
\hline
D3FR &87.63 & 66.50 & 81.10 \\
TDDFA & 63.45 & 55.49 & 58.16 \\
MGCNet &80.87 & 66.01 & 72.36 \\
RingNet &80.05 & 69.46& 72.40 \\
Ours & \textbf{94.47} & \textbf{80.78}  & \textbf{86.40} \\
\hline

\end{tabular}
\end{center}
\caption{Face verification accuracy (\%) on the LFW, CFP-FP and YTF datasets. Our results are obtained using the  weighted center loss. We compare our results with 3DMM-CNN\cite{tuan2017regressing}, Liu et al.\cite{liu2018disentangling}, D3FR\cite{deng2019accurate}, TDDFA\cite{guo2020towards}, MGCNet\cite{shang2020self} and RingNet\cite{sanyal2019learning}.}
\label{tab:FG_acc2}
\end{table} 

\begin{figure}[t]
\begin{center}
\centering 
\includegraphics[width=9cm,trim={1cm 7cm 0cm 8.2cm},clip,angle=0]{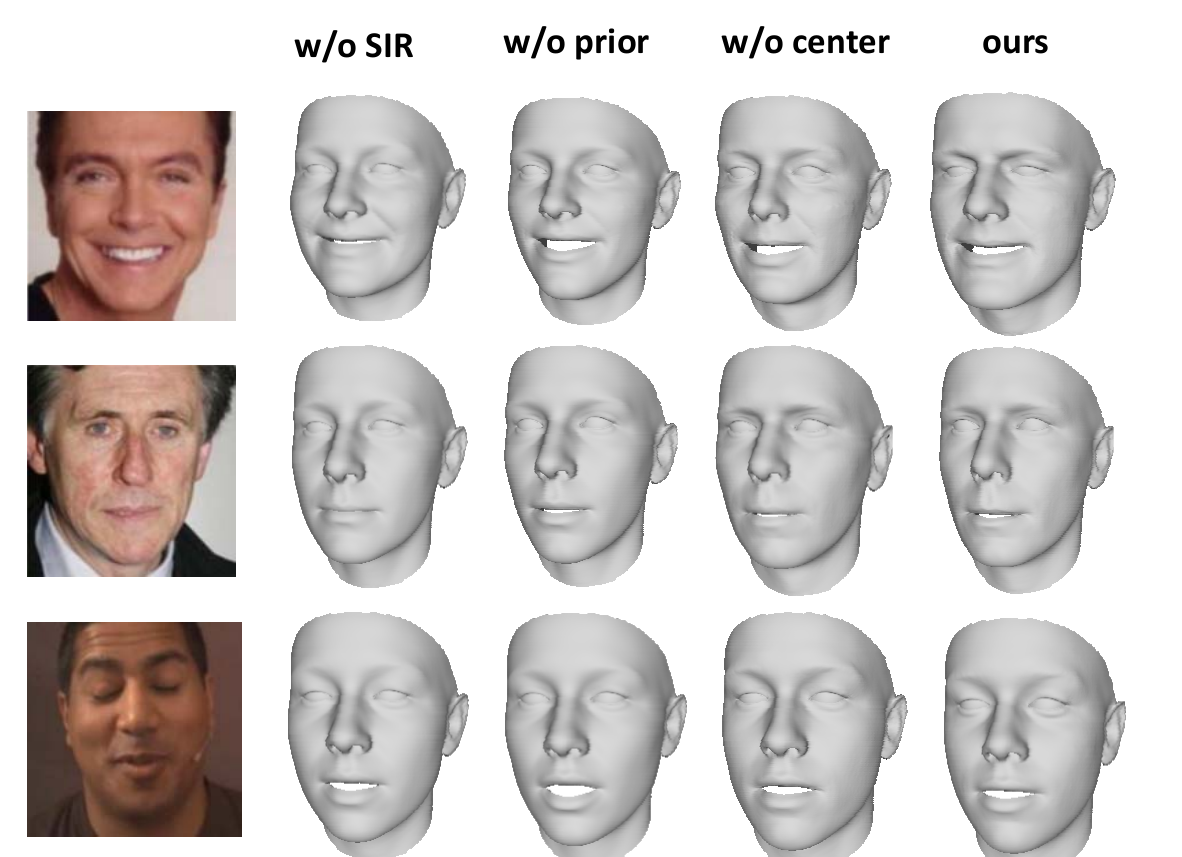} 
\end{center}   
\caption{The ablation study of SIR loss terms.  `w/o' SIR means that we do not use the SIR term in training. `w/o' prior means that we do not use the KL loss in training. `w/o' center means that we do not use the weighted center loss in training. }
\label{fig:v_ablation}
\end{figure}

\begin{figure}[t]
\begin{center}
\centering 
\includegraphics[width=9cm,trim={2cm 3.5cm 9cm 3cm},clip,angle=0]{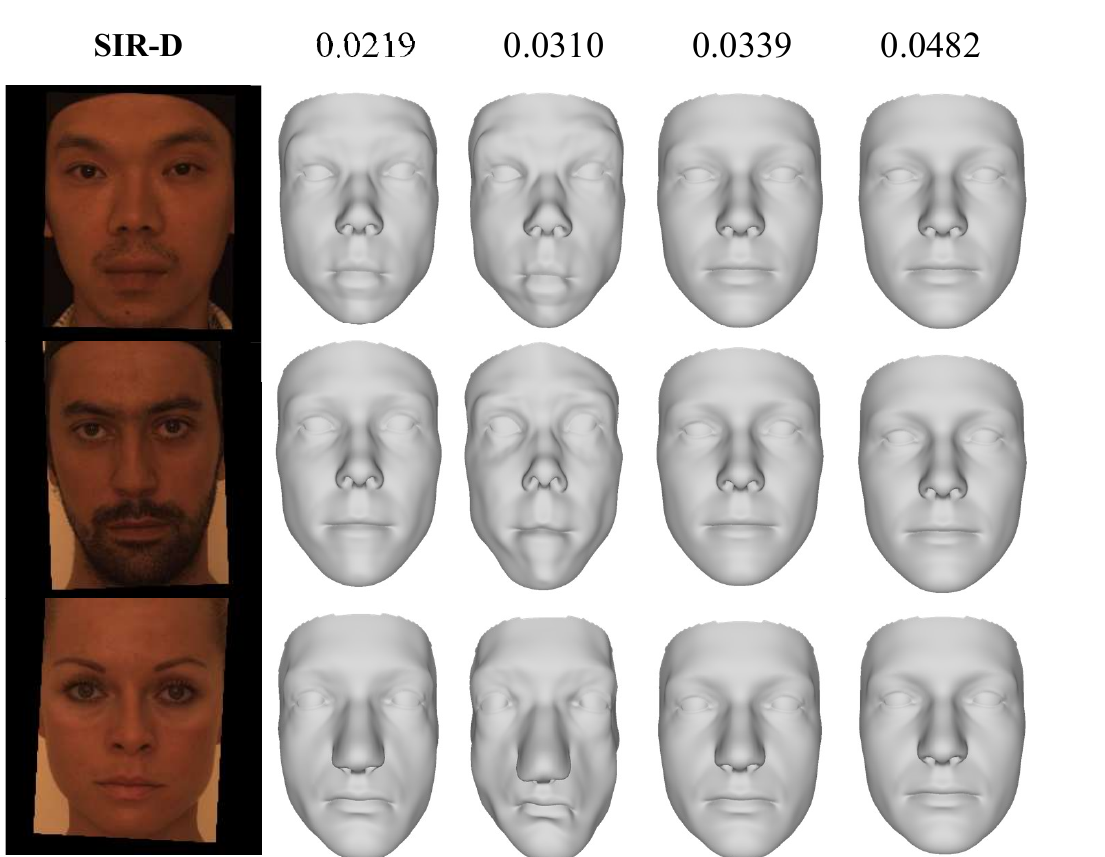} 
\end{center}   
\caption{The samples of the user study on evaluating the SIR-D metric. We present \yl{four} images with different SIR-D values and ask the participant to select the face shape which is most like the input image.}
\label{fig:sird}
\end{figure}

\begin{figure}[t]
\begin{center}
\centering 
\includegraphics[width=9cm,trim={2cm 2.5cm 9cm 2cm},clip,angle=0]{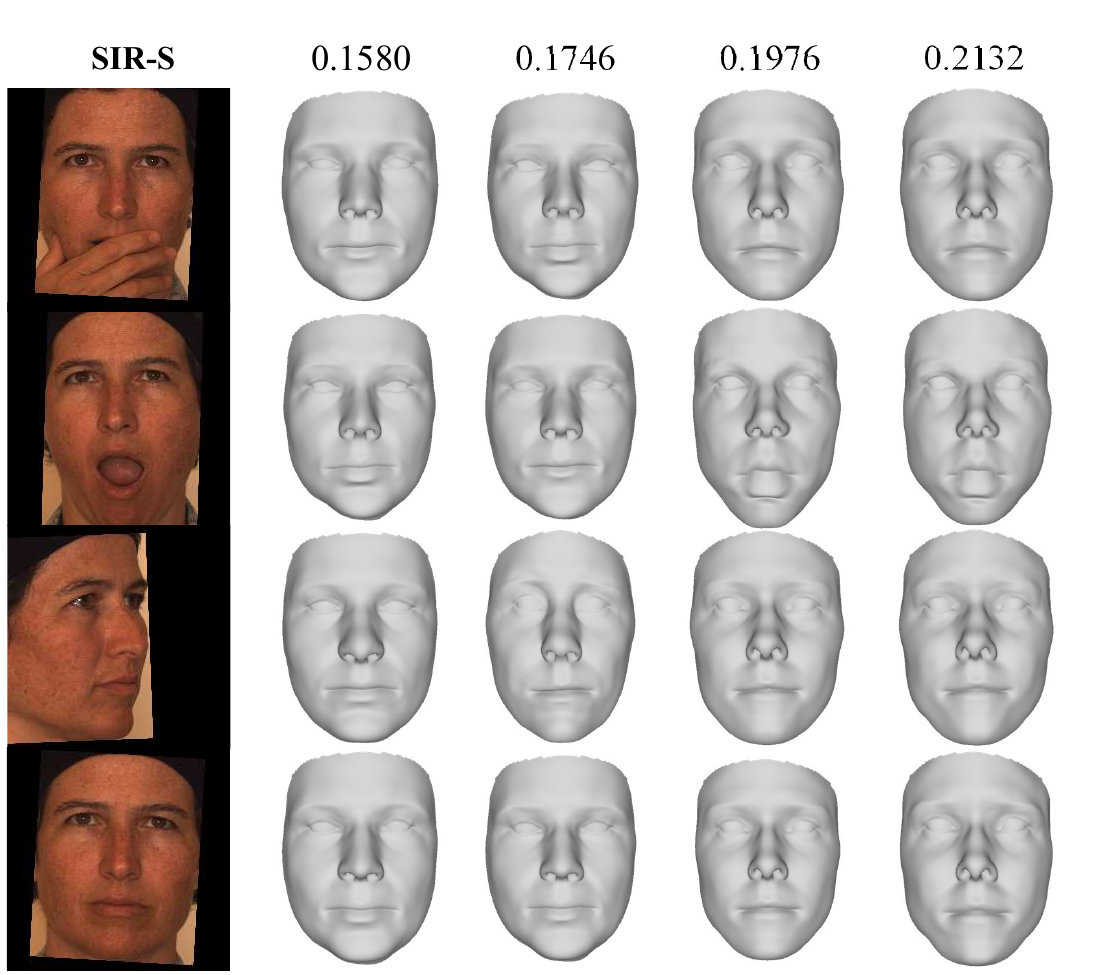} 
\end{center}   
\caption{The samples of the user study on evaluating the SIR-S metric. We present \yl{four} images with different SIR-S values and ask the participant to select the face shape which is most like the input neutral frontal image.}
\label{fig:sirs}
\end{figure}

\textbf{Results on testing benchmarks.}
As mentioned above, the shape parameters must be able to minimize the intraclass distance and maximize the interclass distance in Euclidean space. To evaluate the effectiveness of SIR loss when learning the discriminative shape parameters, we test the face recognition performance of the 3DMM shape parameters on LFW, CFP-FP and YTF.

\textbf{Ablation study.} To validate the efficiency of each loss in our proposed SIR term, we test face verification under various loss combinations. As shown in Table \ref{tab:FG_acc}, the face verification accuracy is very low without any SIR loss terms. When we \YL{add the} center loss in addition to the cosloss, the accuracy increases significantly because our evaluation method is based on the Euclidean distance between parameters, while the center loss can reduce the Euclidean distance between parameters belonging to the same class. The difference in face verification accuracy between the center loss and weighted center loss is subtle because the weighted center loss \yl{only makes} the class center closer to its neutral frontal face parameters. It updates the center by assigning higher weights to the neutral frontal face parameters. This operation does not significantly impact the effect of face recognition but benefits face reconstruction, as shown in Table \ref{tab:FS_error}. 

\begin{table}
\begin{center}
\begin{tabular}{lccc}
\hline
\toprule
Representation & LFW &  CFP  & YTF \\
\midrule 
Parameter  & 94.47  & 80.73 & 86.40\\

Vertices & 94.72  & 80.71 & 86.40 \\
\bottomrule 

\end{tabular}
\end{center}
\caption{The results of face verification (\%) using different identity representations. We use shape parameters and shape vertices as the identity representation.} 
\label{tab:FR1}
\end{table}

Table \ref{tab:FG_acc2} shows a comparison between our results and those of other methods on LFW, CFP-FP and YTF. Note that the other methods may use the cosine distance to measure the parameter similarity. However, the Euclidean distance between parameters can better reflect the difference between geometries and is thus more appropriate to use Euclidean distance. For a fair comparison, we also show the result of our method with the cosine distance.

To demonstrate that the SIR loss can transfer parameter discrimination to geometric discrimination, we compare face verification \YL{performance} on LFW, CFP and YTF with the shape parameters and reconstructed vertices as the identity representation. Table \ref{tab:FR1} shows that the discriminative property successfully transfers from the parameter space to the geometric space. However, we use the BFM model provided by 3DDFA, which deletes some vertices on the original BFM model from PCA. Thus, the shape basis is not strictly orthonormal and the face verification accuracy of vertices has \yl{slight} difference with the \YL{shape parameters}.
\begin{figure}[t]
\begin{center}
\centering 
\includegraphics[width=9cm,trim={0.5cm 3cm 9cm 3cm},clip,angle=0]{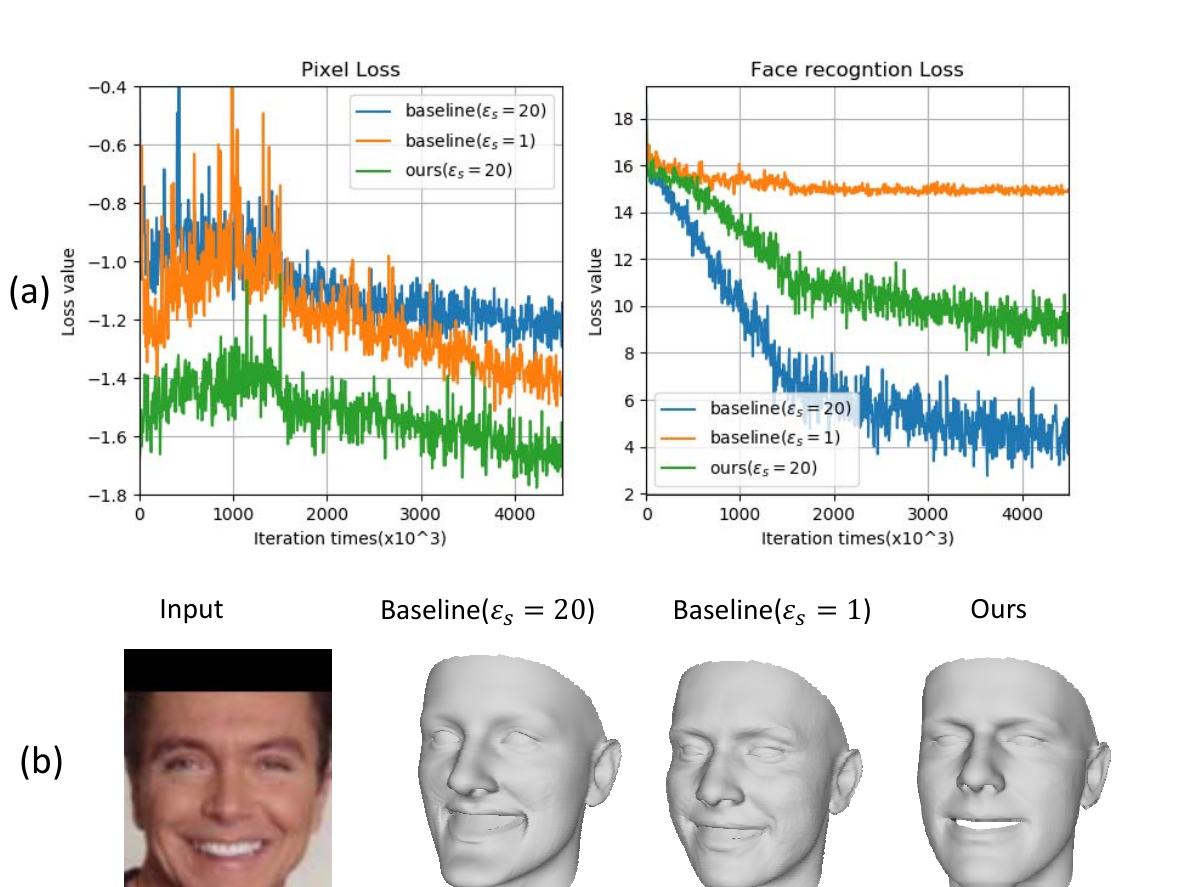} 
\end{center}   
\caption{Comparisons to baseline models for feature separation and training convergence. (a) shows that how the pixel loss(Equation\eqref{pixelLoss}) and face recognition loss(Equation\eqref{Eq:Lrecog}) change with the number of training iterations in the second stage of training. $\varepsilon_{s}$ is used in Equation \eqref{loss_overall}. The weights of the other losses remain the same. (b) shows the visualized reconstructed faces.}
\label{fig:v_loss}
\end{figure}

\begin{figure*}[t]
\begin{center}
\centering 
\includegraphics[width=16cm,trim={0cm 4.5cm 0cm 5.7cm},clip,angle=0]{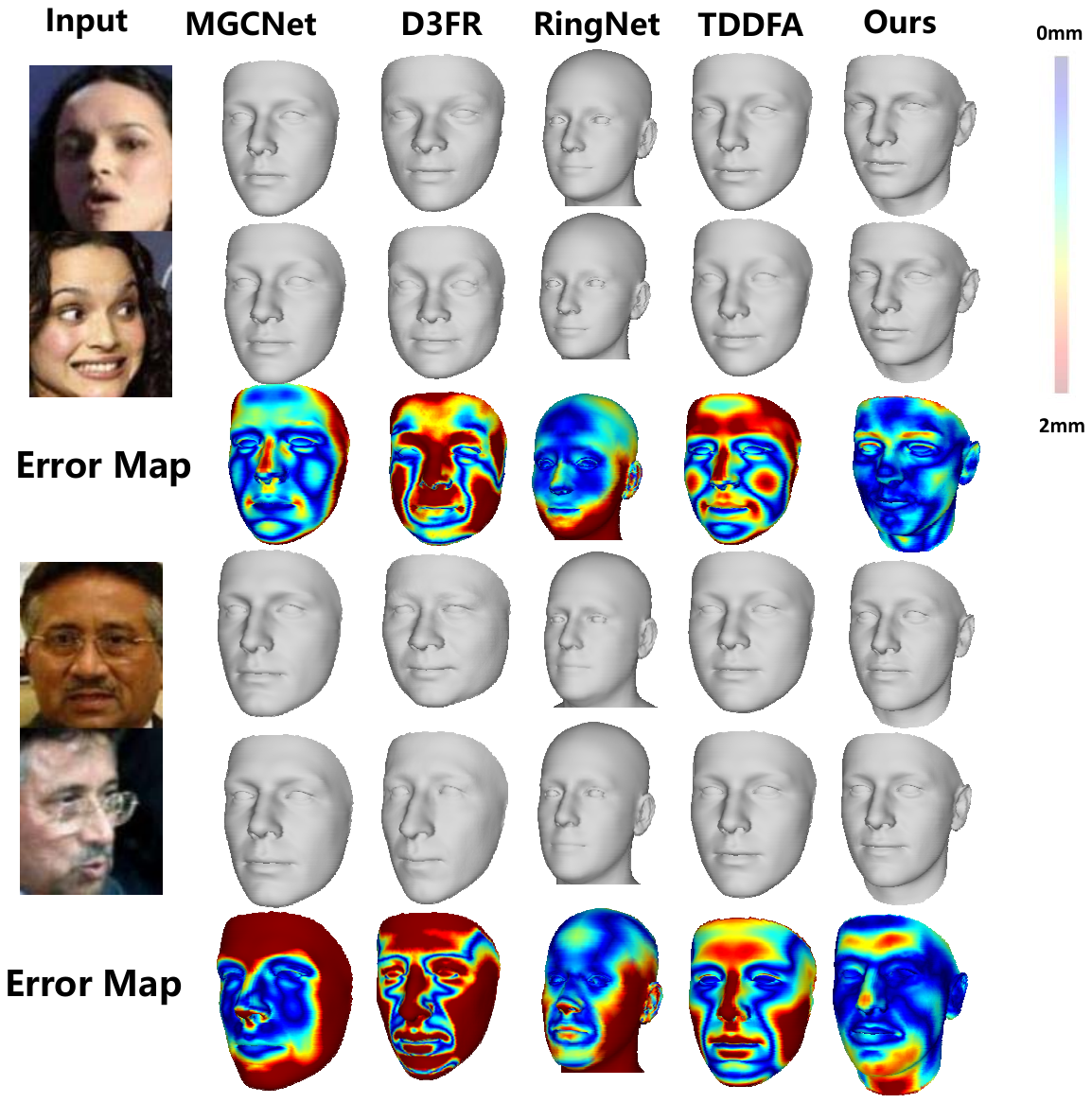}

\end{center}   
\caption{Comparison of our qualitative results under various levels of illumination, various facial expressions, large poses, and occlusion with  MGCNet\cite{shang2020self}, D3FR\cite{deng2019accurate}, TDDFA\cite{guo2020towards} and RingNet\cite{sanyal2019learning} on the LFW dataset. We use only shape parameters to reconstruct the face geometries; thus, normalization occurs without expression and pose effects. The error maps reveal the Euclidean distance between two \YL{shapes}.
}
\label{fig:CFP_V}
\end{figure*}

\begin{figure*}[t]
\begin{center}   
\centering 
\includegraphics[width=16cm,trim={0cm, 8cm, 0cm 8cm},clip,angle=0]{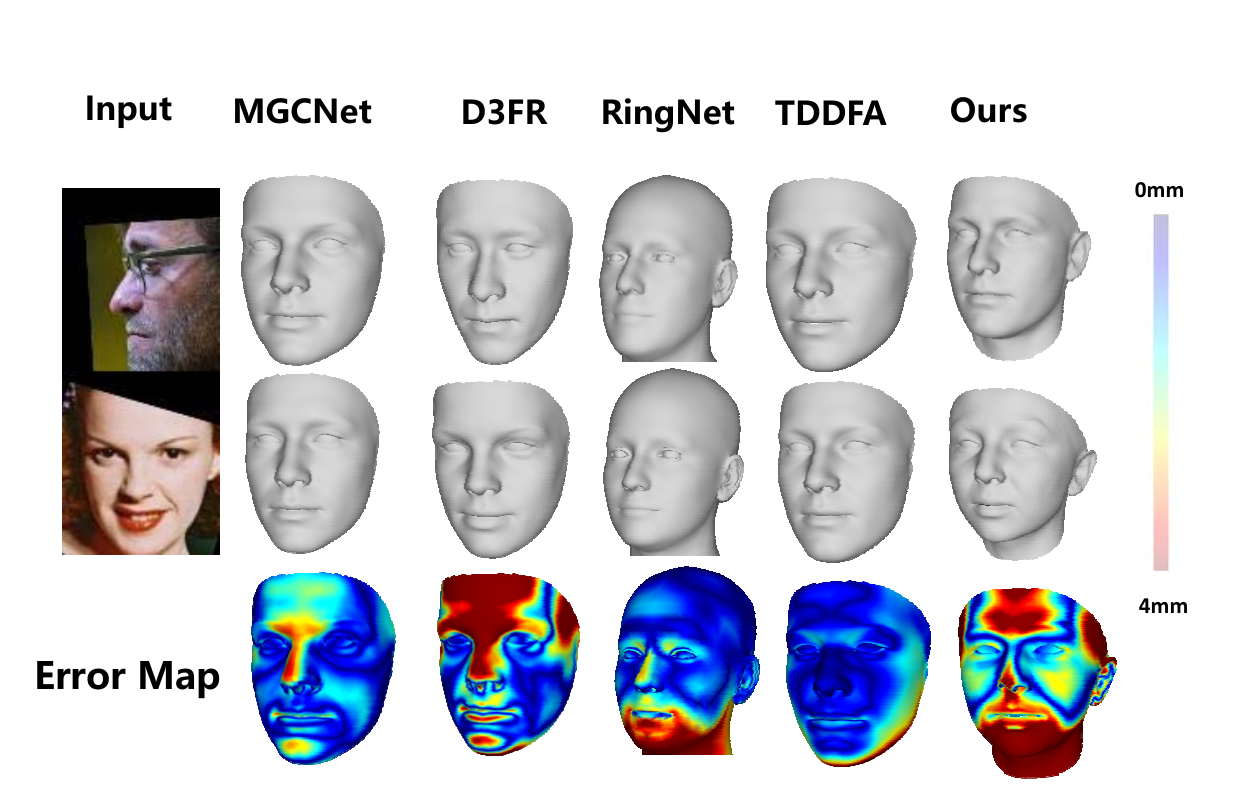} 

\end{center}   
\caption{Comparison of differences between two face shapes, which are regressed from images of different people. The last row presents the error maps, which reveal the difference between the two face shapes by the MGCNet\cite{shang2020self}, D3FR\cite{deng2019accurate}, TDDFA\cite{guo2020towards} and RingNet\cite{sanyal2019learning}. We use only shape parameters to reconstruct the face geometries; thus, normalization occurs without expressions and pose effects.
}
\label{fig:CFP_V2} 
\end{figure*}

\begin{table}
\begin{center}
\begin{tabular}{|l|c|c|c|c|}
\hline

Method &Tran et al. &Lui et al.&MGCNet &D3FR \\
\hline
RMSE   & 2.27 & 2.00 & 1.94  & 1.82\\
\hline
Method &RingNet & TDDFA & Ours-lc & Ours-lwc\\
\hline
RMSE   & 1.84 & 2.01 & 1.82  &\textbf{1.80}\\
\hline 

\end{tabular}
\end{center}
\caption{The face reconstruction error of Florence dataset. We compare our method with Tran et al.\cite{tuan2017regressing},  Lui et al.\cite{liu2018disentangling} and MGCnet\cite{shang2020self}, D3FR\cite{deng2019accurate}, TDDFA\cite{guo2020towards} and RingNet\cite{sanyal2019learning}. Our-lc represents that we use center loss and our-lwc represents weighted center loss. }
\label{tab:FS_error}
\end{table}

\begin{table}
\begin{center}
\begin{tabular}{|l|c|c|c|c|}
\hline

SIR-D & M1 & M2 & M3 & M4 \\
\hline
Sorce   & \textbf{8.34} & 5.35 & 4.56  & 2.82\\
\hline
SIR-S & N1 & N2 & N3 & N4 \\
\hline
Sorce   & \textbf{8.84} & 4.01 & 2.84  & 2.80\\
\hline 

\end{tabular}
\end{center}
\caption{The result of the user study on SIR-scores. M1-M4 are the models with different SIR-D values from low to high. And N1-N4 are the models with different SIR-S values from low to high. These scores are scored by the participants based on the recognizability of the face shapes.}
\label{tab:sir_c}
\end{table}

\begin{table}
\begin{center}
\begin{tabular}{|l|c|c|c|c|}
\hline

Method & TDDFA & MGCNet & D3FR & Ours \\
\hline
SIR-D   & 0.0194 & 0.0465 &  0.0353  &\textbf{0.0049}\\
\hline
SIR-S &  0.1758 & 0.1734 & 0 0.1661  &  \textbf{0.1619} \\
\hline

\end{tabular}
\end{center}
\caption{The SIR score result of MGCNet\cite{shang2020self}, D3FR\cite{deng2019accurate}, TDDFA\cite{guo2020towards} and ours.}
\label{tab:sir}
\end{table}

\begin{figure*}[htbp]
\begin{center}   
\centering 
\includegraphics[width=18cm,trim={6cm 5cm 4cm 1cm},clip,angle=0]{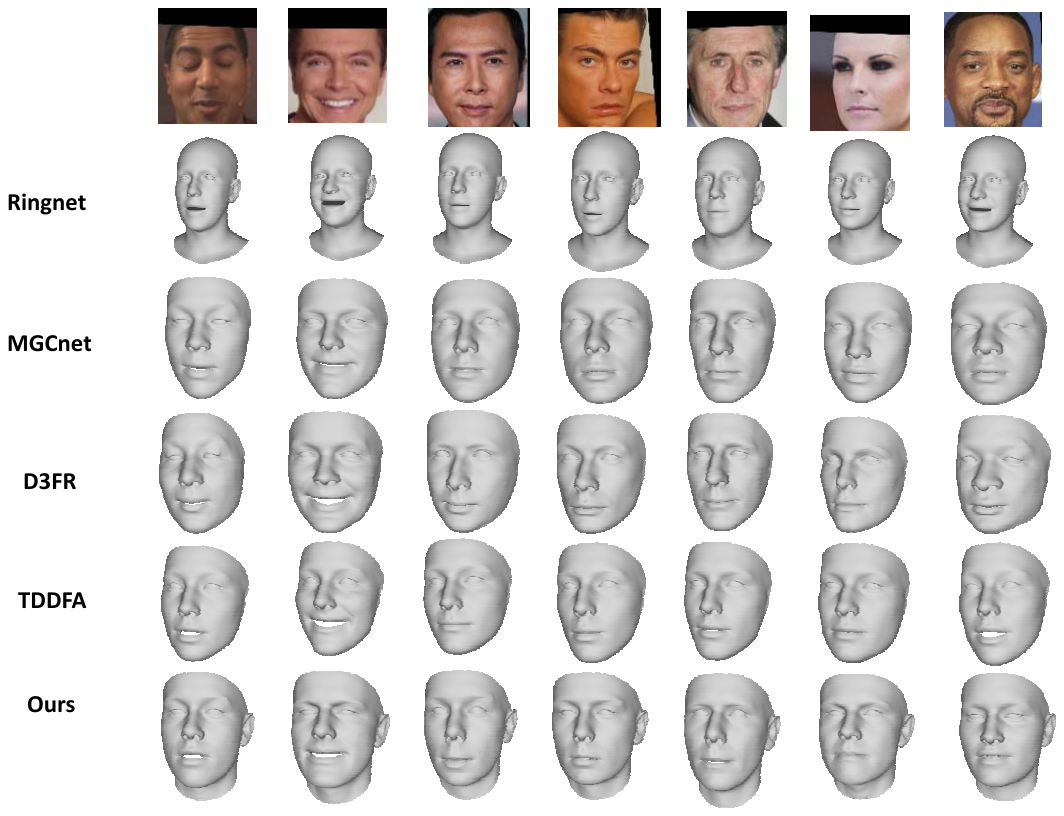} 

\end{center}   
\caption{Comparison of our qualitative results with  RingNet\cite{sanyal2019learning}, MGCNet\cite{shang2020self}, D3FR\cite{deng2019accurate} and TDDFA\cite{guo2020towards}.
}
\label{fig:MICC_V} 
\end{figure*}
\jdq{
\subsection{SIR-Score}
The SIR-Score metric measures the recognizability of face shapes from the following two aspects: distinguishability   and separation, respectively called SIR-D, SIR-S. The SIR-D reveals the distinguishability of the reconstructed face shapes, measured by KL divergence:
\begin{equation}
    L_{SIR-D} = KL(\boldsymbol{P}(\boldsymbol{\alpha}
/ \boldsymbol{\sigma} \mid \boldsymbol{I}, \boldsymbol{\theta}) \|
\boldsymbol{N}(0,1))
\label{Eq:SIRD}
\end{equation}
where $\alpha$ are shape parameters, $\sigma$ is shape eigenvalue, $I$ is the input image and $\theta$ represents the network parameters. A small SIR-D leads to face shape parameters with the same distribution as the training dataset. The SIR-S reveals the separation of reconstructed face shapes, which is the ratio of the inner-class Euclidean distance to the inter-class Euclidean distance:

\begin{equation}
\begin{aligned}
     & \mathbf{W} = \frac{1}{n}\sum^k_{q=1}\sum_{x \in C_q}(x-c_q)(x-c_q)^T \\
     & \mathbf{B} =  \frac{1}{n}\sum_{i\neq j}(c_i-c_j)^T \\
     & L_{SIR-S} =    \mathbf{W}/\mathbf{B}
\end{aligned}
\end{equation}
where $C_q$ is the set of points of the class $q$, $c_q$ is the shape parameters regressed by its neutral frontal face of the class $q$, $n_q$ is the number of points of the class $q$, $k$ is the number of classes and $n$ is the number of samples.

To evaluate the effectiveness of the SIR-Scores when measuring the reconizability of face shapes reconstructed  from shape parameters, we conducted two user studies on SIR-D and SIR-S respectively. 
10 participants joined our user study. The participants are five men and five women, ranging in age from 20 to 61 years old.  We calculated the SIR-Scores on Bosphorus dataset\cite{savran2008bosphorus} and presented four reconstructed 3D models of one person using different SIR-D values and a fixed SIR-S value to the participants to evaluate the SIR-D metric. Figure \ref{fig:sird} shows the samples we used for evaluating SIR-D. Each row presents the four face shape with different SIR-D values. Since SIR-D reflects the similarity between the reconstructed shape parameter distribution and the prior distribution, we find that smaller differences between face shapes produce larger SIR-D values and face shapes with the smallest SIR-D value has the best visual result.  For generating models with a fixed SIR-S value, we multiplied various factors to the shape parameters. 
We then show models of 100 different people and ask participants to give each model a score from 0 to 10, where 10 means its identity is the easiest to be recognizable visually, and 0 means the hardest. Table \ref{tab:sir_c} reports the mean score and shows that smaller SIR-D values means a better discriminativity.
Figure \ref{fig:sirs} shows the samples of our user study on the effect of SIR-S. Different rows are for the same people with different poses and expressions. The last row is the neutral frontal face image. We present the four models with different SIR-S values, where the SIR-D values are similar.  We find that the reconstructed face shapes are more stable and closer to the meshes regressed from the neutral frontal face image with smaller SIR-S values. In this user study, we generated samples of 20 different people, where each person has ten different poses and expressions. We asked participants to rate each mesh model by a score from 0 to 10, where higher scores means that the identities of reconstructed face shapes are the more stable for the different images of the same person, and the face shapes are closer to the shapes regressed from the neutral frontal face image. Table \ref{tab:sir_c} reports the mean scores for each model and shows that smaller SIR-S values means a better stability. Table \ref{tab:sir} reports SIR-scores for all the tested methods, where ours is the best.
}

\begin{table}
\begin{center}
\begin{tabular}{|l|c|c|c|c|}
\hline

Method & TDDFA & MGCNet & D3FR & Ours \\
\hline
SIR-D   & 0.747  & 0.677 &  0.696  &\textbf{ 0.640}\\

\hline

\end{tabular}
\end{center}
\caption{The quantitative result in LFW of MGCNet\cite{shang2020self}, D3FR\cite{deng2019accurate}, TDDFA\cite{guo2020towards} and ours.}
\label{tab:vist}
\end{table}

\subsection{Quantitative Results on Shape Reconstruction}
To evaluate the stability of our algorithm and the precision of the reconstructed 3D face shapes, we calculate the RMSE between neutral frontal 3D scans and 3D shape faces regressed from images under various conditions, including illumination, head pose, expression, and occlusion, on the Florence dataset\cite{bagdanov2011florence}.

\textbf{Test on the Florence dataset.} The MICC dataset contains the 2D/3D faces of 53 subjects, including two indoor videos, one outdoor video and the faces' 3D models. The 3D face model of each person includes one or two frontal faces with a neutral expression. Unconstrained outdoor videos are recorded under natural lighting conditions, which are more challenging. In our experiment, we choose the outdoor video frames as the input and randomly select 100 frames of each subject to form a test dataset that contains 5,300 face images. In our preprocessing stage, the face and its landmarks are detected by the MTCNN\cite{zhang2016joint}. Afterward, the faces are aligned using similarity transformation and cropped to 112$\times$96 in the RGB format. The ground-truth scans are cropped at a radius of 95 mm around the nose tip, and the meshes generated are aligned to the ground truth using ICP with an isotropic scale. 

When testing, we reconstruct only the shape of the face without expressions. Table \ref{tab:FS_error} reports the RMSE of the point-to-plane distance between the ground-truth face shape and reconstructed face shape after ICP on an isotropic scale. Figure \ref{fig:CED_fig} shows the cumulative error distribution (CED) curves of different methods.

As shown in Table \ref{tab:FS_error}, the reconstruction error of the weighted center loss is lower than that of the center loss. The reason is that the weighted center loss pushes the regressed shape \YL{parameters} to the values regressed from the corresponding neutral frontal face image. The parameters obtained from neutral frontal face images are more accurate than those of profiles and faces with extreme expressions. Therefore, using neutral frontal face parameters as the class center can improve the accuracy of reconstruction accordingly. The RMSE of our method on the Florence dataset is also lower than that of other state-of-the-art methods. Note that some results are inconsistent with the results reported in their paper, since they ran their methods on each frame of Florence's videos and averaged each video's results to obtain a single reconstruction. Since the averaging operation prevents the results from showing the reconstruction stability, we evaluate those methods using the same approach as presented above.

\subsection{Qualitative Results on Shape Reconstruction}

\textbf{Ablation Study.}
Figure \ref{fig:v_ablation} shows reconstructed face shapes with different combinations of losses. As presented, some identity details are missing without the SIR item. If we use identity loss without a distribution prior, the identity discrimination in the parameter space cannot be effectually transferred to the appearance space since the norm the shape parameters would be small, leading to the reconstruction of a mean face. The weighted center loss aggregates shape parameters to the frontal face in Euclidean space. It improves the reconstruction robustness to different face poses and facial expressions. The last column of Figure \ref{fig:v_ablation} shows that the weighted center loss helps strengthen the identity discrimination.

We compare our method with the baseline model, in which the original structure of SphereFace\cite{liu2017sphereface} is utilized. As shown in Figure \ref{fig:v_loss}, the face recognition loss of the baseline model drops faster than ours using the same loss weights, but their pixel loss is much higher than ours. The visualization results of the reconstructed faces are not satisfactory. If we reduce the weight of the SIR loss to improve the face reconstruction performance, the face recognition loss of the baseline model does not converge, and the reconstructed face recognizability is also weakened.

To evaluate our reconstructed face's stability and visual identifiability under challenging conditions such as various types of illumination, large poses, various expressions and occlusion, we compare the qualitative results of estimating the face shape from a single image in the LFW database obtained with MGCNet\cite{shang2020self}, D3FR\cite{deng2019accurate}, TDDFA\cite{guo2020towards} and RingNet\cite{sanyal2019learning}. As Figure \ref{fig:CFP_V} shows, we choose images under four conditions: various levels of illumination, various facial expressions, large poses and occlusion.

\textbf{Occlusion.} As shown at the top of Figure \ref{fig:CFP_V}, the woman's hair occludes her cheeks. Thus, we cannot directly infer the shape around the cheeks from the picture. Therefore, different orbital geometries can easily be regressed if the constraints of face recognition are not used, such as the result of MGCNet\cite{shang2020self}. The SIR term can aggregate the same person's features and infer the geometric information of the occluded part.

\textbf{Expression.}
The final geometry is affected by both of the face shape and its expression, which means that the same face geometry can be determined by different combinations of face shapes and expressions. The same person with different expressions may regress varying face shapes. As shown at the top of Figure \ref{fig:CFP_V}, for the case of smiling, the other methods have some errors in the mouth area, while our reconstruction results remain stable.

\textbf{Pose.}
Large poses result in some information loss regarding the face shape due to self-occlusion. As shown at the bottom of Figure \ref{fig:CFP_V}, the face contour is difficult to estimate. However, the SIR term can push the regression parameters from the profile to the parameters regressed from the frontal face and infer the missing information.

\textbf{Illumination.}
As shown in the top row of Figure \ref{fig:CFP_V}, the contour of a face could be unclear due to inappropriate illumination. Similar effects could be caused by some special poses or occlusion, where some shape information is lost. SIR loss can infer the lost information for the reason described above.

\jdq{We also conduct the quantitative evaluation on the LFW dataset using the ratio of RMSE between pairs of images with the same identity and between pairs with different identities:
\begin{equation}
\begin{aligned}
     & \mathbf{W} =  \frac{1}{n_s} \sum_{i \in C_s}| \left | S_{i1} - S_{i2} | \right | \\
     & \mathbf{B} =   \frac{1}{n_d} \sum_{j \in C_d} | \left | S_{j1} - S_{j2} | \right | \\
     & L =  \mathbf{W}/\mathbf{B}
\end{aligned}
\end{equation}
where $C_s$ is the pair of the same identity pair in LFW test benchmark and $C_d$ is the pair with different identities. $S*$ is the reconstructed face shape. $n_s$ is the number of the pairs with the same identity and  $n_d$ is the number of the pairs with different identities. Table \ref{tab:vist} shows that our method has the best performance.

}

In addition, the reconstructed face shapes of the same person in different environments should be the same. The face shapes reconstructed for different people, however, need to differ from each other. Figure \ref{fig:CFP_V2} shows the difference in the reconstructed shapes from different people. It shows that the face shapes reconstructed by other methods are similar. In contrast, our result presents the expected differences.

Figure \ref{fig:MICC_V} shows a qualitative comparison between our method and other state-of-the-art methods. Different from  RingNet\cite{sanyal2019learning}, TDDFA\cite{guo2020towards}, our results maintain the identity feature of rhe input images. MGCNet\cite{shang2020self} and D3FR\cite{deng2019accurate} use photometric metrics and can effectively capture the identity feature of the input image. However, they produce poor results when the faces have extreme expressions and large poses as shown in the second and fourth columns in Figure \ref{fig:MICC_V}). In contrast, our method can capture identity features under challenging conditions due to the benefits from our identity losses. 

\section{Conclusions}


Our research started from the observation that despite the high face recognition accuracy obtained using the 3DMM shape parameter, the reconstructed 3D face shapes are lack of significant visual discrimination. We first explored the relationship between the between 3DMM parameter space and 3D geometric space, and propose SIR losses that explicitly enforce shape consistency in the shape parameter space while implicitly guiding reconstructed face shapes to be visually discriminative. In detail, the identification loss explicitly maximizes the interclass and minimizes the intraclass Euclidean distance of shape parameters while it implicitly maximizes/minimizes the MSE of the shape geometry of different people/the same person. Kullback–Leibler losses are also utilized to explicitly constrain the shape parameters to follow a particular distribution and implicitly let them to share the same visual distinction as the shapes used to train the 3DMMs. We build a neural network and an associated training strategy to cope with the lack of such a dataset that contains both identity and 3D geometry annotations, which can quickly converge under our training strategy. \jdq{Finally, we propose the SIR-score metric to evaluate the recognizability of face shapes.} The experiments show that our results outperform those of the state-of-the-art methods in terms of the reconstruction error, visual discrimination, and face recognition accuracy.
\bibliographystyle{eg-alpha-doi} 
\bibliography{egbibsample}       



\end{document}